# Implementing Online Reinforcement Learning
## *with*
## Temporal Neural Networks

J. E. Smith

University of Wisconsin-Madison (Emeritus)

01/29/2022

## 1. Introduction

Temporal neural networks (TNNs) encode information as precise spike timing relationships. The term "TNN" is used here to distinguish networks that use temporal communication and computation from today's broader class of spiking neural networks (SNNs) many of which use spike *rates* for communication and computation, rather than *individual spike timing relationships*. Moreover, the TNNs considered here employ online localized learning via spike timing dependency (STDP) to support continual, adaptive online learning. This contrasts with a large fraction of proposed TNNs that rely on compute-intensive back propagation learning methods, adapted, transferred, or inspired by conventional artificial neural networks (ANNs).

Of the TNNs that provide STDP-like learning, most have been developed for offline classification problems -- as exemplified by the MNIST benchmark [1] [6][8][9][11][16][17]. Although offline supervised classification benchmarks are useful for research studies, as far as practical applications these types of problems do not play to the strengths of TNNs with STDP learning. As opinion: the likelihood that any of the proposed TNNs or SNNs will replace deep convolutional ANNs for offline classification problems is extremely small. For those types of problems, deep convolutional neural networks have been thoroughly refined, are highly successful, and are widely used (albeit with extremely compute-intensive offline training methods). The future of TNNs and STDP lies in efficient implementation of online applications, especially those that employ online reinforcement learning (RL).

In this document, a TNN architecture for implementing efficient online RL is proposed and studied via simulation. The proposed *T-learning* system is composed of a frontend TNN that implements *online unsupervised clustering* and a backend TNN that implements *online reinforcement learning*. The reinforcement learning paradigm employs biologically plausible neo-Hebbian three-factor learning rules as articulated by Gerstner et al. [4]. As a working example, a prototype implementation of the cart-pole problem (balancing an inverted pendulum) is studied via simulation.

## 2. Background

A generic reinforcement learning system is illustrated in Figure 1. The system observes a sequence of inputs in the form of *state variables* coming from the *Environment*. An *Agent* takes the combination of input state variables and *Learned State* to enact a *Policy* that produces an *action* which is then applied to the environment. In response to actions, and possibly other external environmental variables (not shown), the *Environment* produces new state variables and, optionally, a *reward* (a *punishment* is a negative reward). The new state variables and accompanying reward (if any) are then assimilated into the *Learned State*.

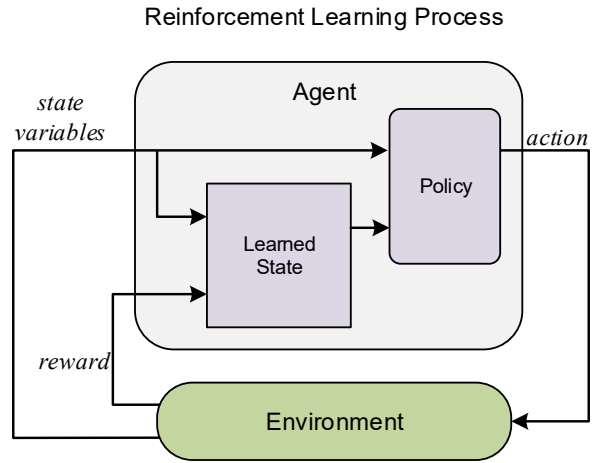

**Figure 1. Generic RL System Architecture.**

In *online* systems as considered here, learning operates continually as a sequence of observed state variables and rewards are presented to the inputs. For each observation, the learning method updates *Learned State* (typically embodied as weights), and then moves on to the next observation.

## 2.1 Q-Learning Baseline

Q-Learning [15] serves as a standard for performance comparisons, and Q-learning-based RL implementations provide a good context for motivating and describing T-learning. Q-learning is an off-policy reinforcement learning method -- it strives to learn the optimal action for each input state, irrespective of the policy's actual actions. Q-learning employs a Q-table, which serves as a data structure upon which agent-based policies can be constructed.

Of interest here are systems that traverse sequences of environment states. A state is defined by a set of state variables, whose combined values determine the overall state. A very large number of states may be described with a much smaller number of state variables. For example, 4 state variables with 16 values each yield 64K total states.

The classic Q-table consists of a row for each input state and a column for each output action (see for example Table 1). Each entry in the Q-Table is a relative measure of the expected reward if the given action is taken when observing the given state. Note that the table entries are not probabilities: it is their relative magnitudes that is important. Based on the history of past state traversals and rewards, the Q-table mechanism gradually constructs and incrementally modifies table entries, as described in the next subsection.

**Table 1. Example Q-Table**

|  | | Action ($a$) | | | |
|---|---|---|---|---|---|
|  | Q($s,a$) | 0 | 0 | 2 | ... |
| State ($s$) | 0 | 0.24 | 0.01 | 0.04 | 0.05 |
|  | 1 | 0.02 | 0.02 | 0.1 | 0.04 |
|  | 2 | 0.07 | 0.41 | 0.08 | 0.06 |
|  | 3 | 0.15 | 0.26 | 0.02 | 0.09 |
|  | ... | 0.16 | 0.04 | 0.28 | 0.07 |

## 2.2 Ideal Q-Learning

When in state $s_t$ at time step $t$, some policy -- not necessarily related to current Q-Table contents -- determines an action $a_t$. In response to the action, the environment yields the next state $s_{t+1}$ and a reward $r_{t+1}$. Then, given the 4-tuple $[s_t\ a_t\ s_{t+1}\ r_{t+1}]$, the Q-Table entry at $Q(s_t, a_t)$ is updated according to Bellman's equation:

$$Q(s_t, a_t) \leftarrow Q(s_t, a_t) + \alpha\ (\ r_{t+1} + \gamma\ \max_a Q(s_{t+1}, a) - Q(s_t, a_t))$$

$$= Q(s_t, a_t)(1 - \alpha) + \alpha\ (\ r_{t+1} + \gamma\ \max_a Q(s_{t+1}, a))$$

if $r_{t+1} = 0$, then:

$$= Q(s_t, a_t)(1 - \alpha) + \alpha\gamma\ \max_a Q(s_{t+1}, a)$$

where $\alpha$ is a learning rate and $\gamma$ is a discount factor.

Figure 2 is a block diagram for the classic Q-Learning method. The environment provides state variables as inputs to the system. These are encoded into a one-hot format so there are as many signal lines as there are Q-Table rows. This is input $s_t$ to the Q-Learning mechanism.

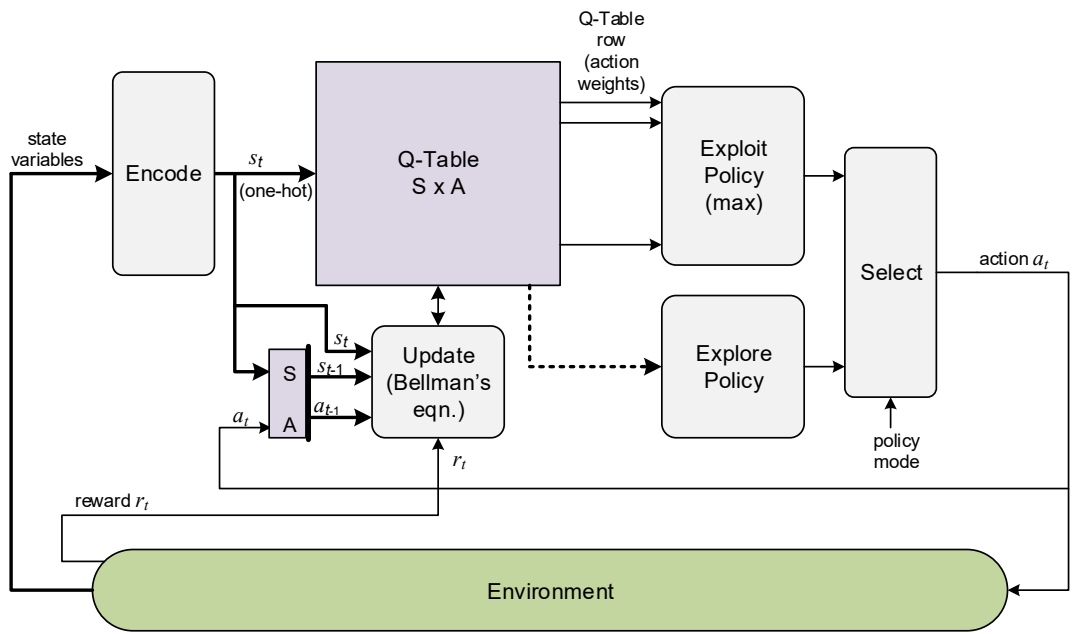

**Figure 2. Classic Q-Table implementation. To be consistent with the T-learning implementation to follow, the state subscripts are time shifted by one unit, i.e. $s_t$ and $s_{t-1}$ are used.**

A policy determines the action to be taken at each step. Initially, when the Q-Table is devoid of information, the policy employs an *exploration* phase. Exploration does not use the Q-Table for making policy decisions. The exploration policy may follow some heuristic or may follow a sequence of pseudo-random actions. At each step, the state and action are buffered ($s_{t-1}$ and $a_{t-1}$ in the figure) to be used for table updating during the next step. After the environment produces the next state and reward, the Q-Table is backward updated via Bellman's equation. The Q-Table is updated at every step regardless of the policy used. By constantly updating with Bellman's equation, rewards are back propagated incrementally throughout the Q-Table, and, in a well-behaved system, entries will converge to a set of values that are optimal or nearly so. After some level of convergence has been achieved, the Q-Learning policy is switched to a phase that *exploits* the Q-Table, i.e., for a given state input, it selects the action associated with the table entry having the highest value. In most implementations the policy occasionally switches back to an exploration phase in an attempt to discover new paths.

Although the classic method works well in theory, in most practical applications the state space (and therefore, the Q-Table) is huge. Not only that, but the convergence process requires many state traversals, and the more states, the more traversals are required. Deep Q-Learning, summarized in the next section is a neural network-based method that deals with the state space problem.

## 2.3 Deep Q-Learning

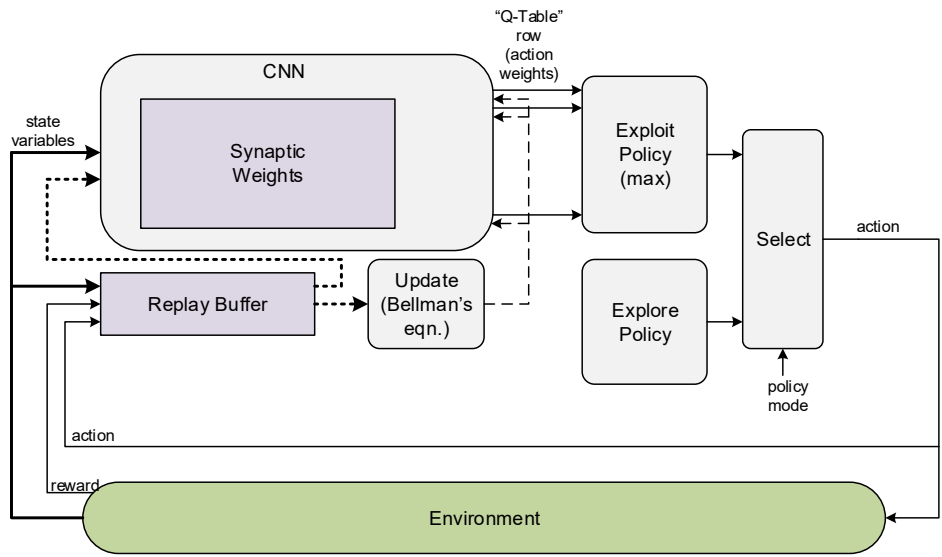

**Figure 3. Deep Q-learning implementation. Dashed lines indicate paths used during training.**

In Deep Q-Learning, a convolutional neural network (CNN) is trained to map the state variables onto what can be described as *approximate* Q-Table rows, and the exponential expansion from state variables to states is avoided. The CNN's weights are updated via a back propagation learning process in which update information is placed in a replay buffer during exploration. After a certain amount of exploration, a version of Bellman's equation is used to determine "correct" outputs for a given set of inputs, and the correct values are used as part of a brief offline back propagation learning session. These offline learning sessions are intermixed with inference. Or, two networks can be used -- one performing inference (and therefore actions), while the other is learning from the replay buffer. Periodically, the networks can be switched.

The CNN that is at the heart of the Deep Q-learning method is trained via conventional back propagation. The wrinkle is that training is done periodically and incrementally by using saved update information held in a replay buffer. Essentially, this is a mini-batch method of training, and although it works, it strikes this author as a rather inelegant semi-offline solution to an online learning problem.

## 2.4 TNN Reinforcement Learning

The overall T-learning architecture is in Figure 4. The design consists of two cascaded TNNs: a Clustering TNN (C-TNN) and a reinforcement learning TNN (R-TNN). The state variables are first presented directly to the C-TNN which performs concurrent inference and unsupervised online learning. In doing so, the C-TNN compresses sets of similar state variables to a single one-hot cluster identifier (CId). The underlying assumption is that *similar* state variables should lead to the *same* action.

The R-TNN uses reinforcement learning to map CIds to actions. This is a one-hot to one-hot mapping, so synaptic learning is simplified in some respects. A reward signal from the environment is broadcast to all the synapses in the R-TNN. When a reward signal is non-zero, a reward (or punishment) is applied, and the reward signal is combined with local spiking history to update the R-TNN's synaptic weights.

Like the C-TNN, the R-TNN operates in an online streaming fashion where inference and training take place continually and concurrently. Although both TNNs are feedforward, this is a recurrent system overall with feedback through the environment.

Observe that the C-TNN follows an off-policy method like the Q-Table. Regardless of the actions taken by the R-TNN, the C-TNN simply clusters its observed input patterns without supervision.

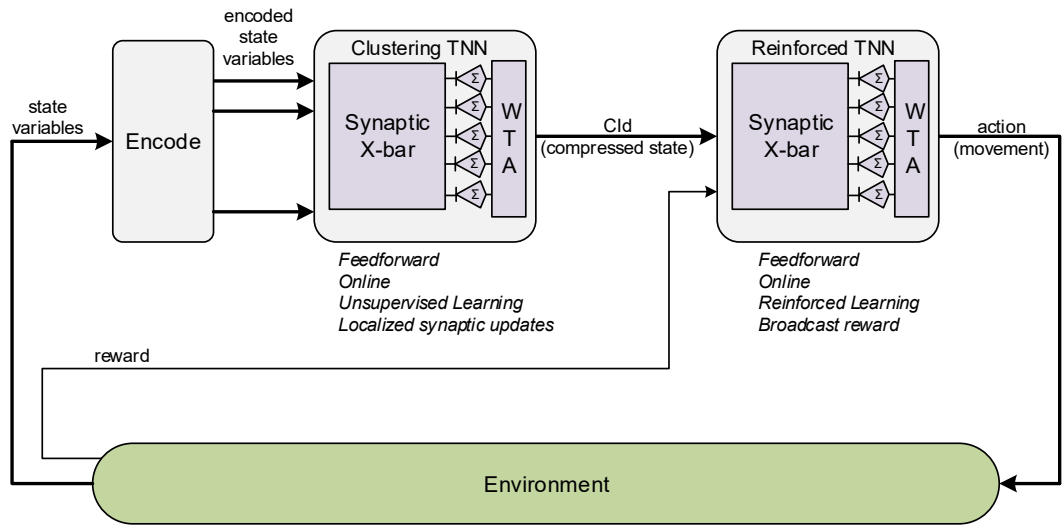

**Figure 4. T-learning system: a TNN implementation of reinforcement learning.**

## 2.5 Credit Assignment

The RL agent implemented as a column produces a sequence of actions. Call this the *action path* or *action sequence*. When there is a reward or punishment, credit assignment determines which of the prior actions should be rewarded (by potentiating associated synapses) or punished (by depressing associated synapses). In credit assignment terms, synaptic updates along the action path are a function of which actions were taken and when.

The three-factor learning rules as described by Gerstner et al. [4] is in this author's opinion the most promising (perhaps the only) biologically plausible credit assignment method. With this approach, the segment of the action path to be updated trails the head of the path (the current action) by a parameterized fixed length. The amount of weight update is maximum at the head of the path and decays along the length of the fixed-length path. Along the path, both synapses that received input spikes and those that did not are potentially updated. Rather than incrementally propagating rewards and punishments as in classic T-learning, all the synaptic weight adjustments occur immediately when there is a reward or punishment.

## 3. T-Learning Architecture

In this section, the T-Learning architecture is described with the cart-pole problem as a running example. The formulation given here uses Nagendra et al. [10] as a starting point -- although it diverges somewhat from that formulation.

## 3.1 Cart-Pole Problem

The objective is to learn to balance an inverted pendulum (a pole) on a moving cart (Figure 5). This is to be done through the actions of applying forces +F and - F to the cart.

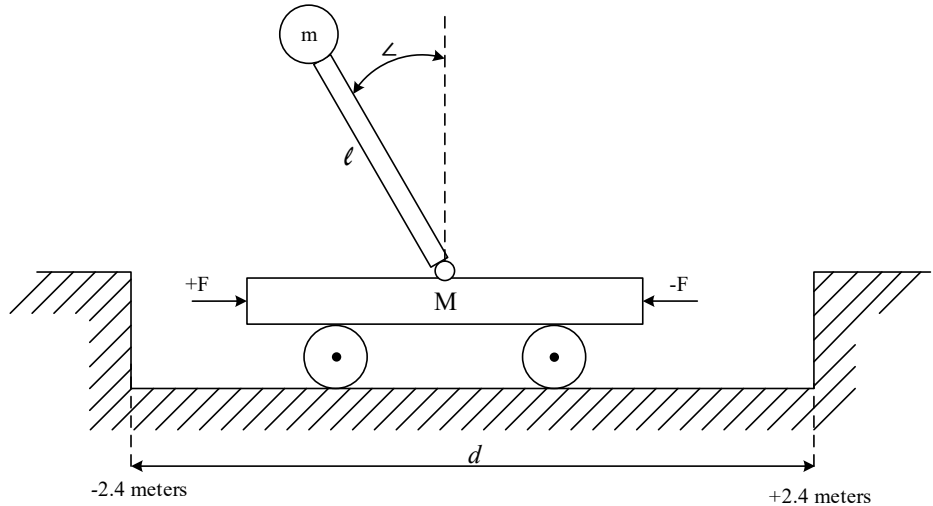

**Figure 5. Cart-Pole dimensions.**

Equations describing the system are given below, taken from [10].

$$\ddot{z} = \{(M+m)g\sin z - \cos z[F+ml\,\dot{z}^2 \sin z\,]\} \,/\, \{(4/3)(M+m)l - ml\cos^2 z\}$$

$$\ddot{d} = \{F+ml[\dot{z}^2\sin z - \ddot{z}\cos z]\} \,/\, (M+m)$$

Parameters for baseline experiments are given below; also taken from [10]

    $M$ = .711 kg                           (1.6 lbs)
    $m$ = .209 kg                           (.46 lbs)
    $g$ = 9.8 m/sec$^2$                  (32 ft/sec$^2$)
    $F$ = ± 10 newtons              (2.25 lbs-force)
    $l$ = .326 meters                  (1.1 ft)
    $\tau$ = .02 sec. time intervals

### 3.2 Implementation Overview

As classically defined, there are four state variables: the angle of the pole $z$ and its first derivative $\dot{z}$, and the cart's displacement $d$ and its first derivative $\dot{d}$.

The model T-learning system is shown in Figure 6. The environment (EV) is a part of the simulation system that models the physics in 64-bit floating point and passes floating point state variables to the encoding block (EN). The Q-Table in the figure is not part of the T-learning system. Rather, it is present in the simulator to allow later performance comparisons with conventional Q-learning. The Q-Table takes the same inputs as the TNN.

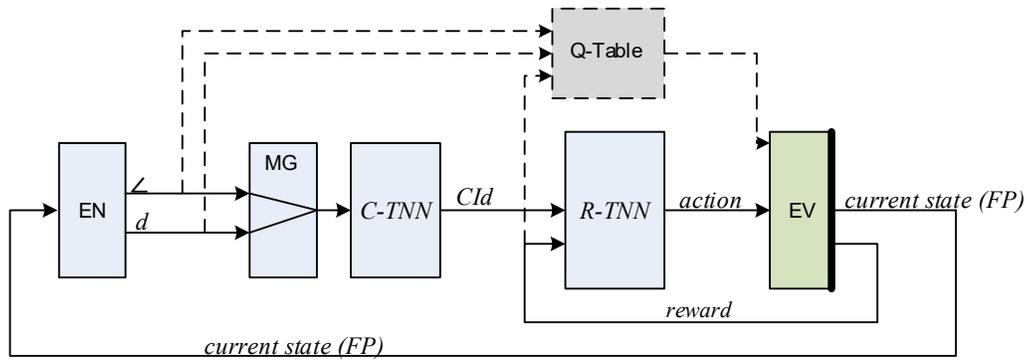

**Figure 6  Simulated T-Learning system. The Q-Table block is in the simulator to provide baseline comparisons; it is not part of the actual TNN system.**

### 3.3  Input Encoding Methods

Because a basic principle behind the RL architecture (Figure 4) is that similar inputs lead to the same action, a critical part of the design process is encoding input information coming from the environment. Given encoded inputs, it is up to the C-TNN to form clusters based on similarity, then all the inputs that belong to the same cluster are treated the same way by the R-TNN. Input volleys may appear similar or dissimilar depending on the chosen encoding method. The type and quality of input encoding is therefore critical for determining the accuracy, efficiency, and hardware cost of TNN processing.

In a sense there are two coding dimensions in spike volley: 1) the presence or absence of a spike on a given line, 2) the time of a spike.  The spike time is low precision to begin with, and binarized volleys may be adequate for many implementations.  Although binarized volleys are used at the inputs, volleys internal to the TNN are not binarized.  For the cart-pole problem input volleys are binarized. If the spike time (strength) dimension is useful for a given application, then it is available to be used.

The encoding process is reduced to encoding an external state variable as a binary volley -- 1-hot volleys in particular.  For example in the cart-pole problem, the pole angle ranges from -12 to +12 degrees that can be divided into 12 equal intervals having endpoints:  [-12, -10, -8, -6, -4, -2, 0, 2, 4, 6, 8, 10, 12]. So, for example the value 10.89 maps to the interval with endpoints [10, 12].

Observe that for this example, not only is the state variable discretized, but some implicit clustering also takes place.  For example the values 10.11, 10.36, 10.89, 11.67 all map to the same interval [10, 12]; i.e., to the same cluster.  When clustering is performed on a single state variable, it is defined to be *unitary clustering*.

Next, observe that when performing unitary clustering over a range of values, the intervals do not have be the same size.  For example, pole angles can be divided into 6 ranges defined by [-12, -6, -1, 0, 1, 6, 12]

Finally, observe that a state variable does not have to span a numerical range.  They can be discrete items such as letters of the alphabet.  In that case, 26 1-hot encoding can be used.  Another option is to use two 1-hot fields, one with 5 lines and the other with 6 to encode the 26 letters.

*Multiple State Variables*

To encode multiple state variables, multiple spike volleys are concatenated, or merged, into a single larger volley. Refer to Figure 7.  First, as part of the encoding process, the range of each external state variable is discretized into $N$ intervals and encoded as  1-hot spike volleys on $N$ lines $x_1$... $x_N$.  $N = 8$ in this example. A value in interval $i$ is encoded with a spike on line $x_i$. In Figure 7a there are three discretized state variables having the values 2 | 3 | 8  .

*Similarity Encoding: m-hot codes*

A 1-hot code is deficient for clustering methods that rely on the number of coincident spikes to determine similarity. With a 1-hot code, two differing input volleys have no coincident spikes, so all encoded values are equally dissimilar.

This deficiency can be resolved by using an *m*-hot code, where *m* is odd. For *N* discrete intervals, the number of lines is extended to *N+m*-1. A value in interval *i* is encoded with spikes on lines $x_i...x_{i+m-1}$. Examples of 3-hot encodings for 2 | 3 | 8 and 2 | 4 | 6 are in Figure 7b and c. Note that values that differ by 1 (3 and 4 in the example) have two spikes in common, reflecting their similarity. Values that differ by 2 (8 and 6 in the example) have one spike in common, reflecting less similarity. Values that differ by more than 2 have no spikes in common indicating dissimilarity. This encoding method is also put forward by Purdy [13] in the context of hierarchical temporal memory.

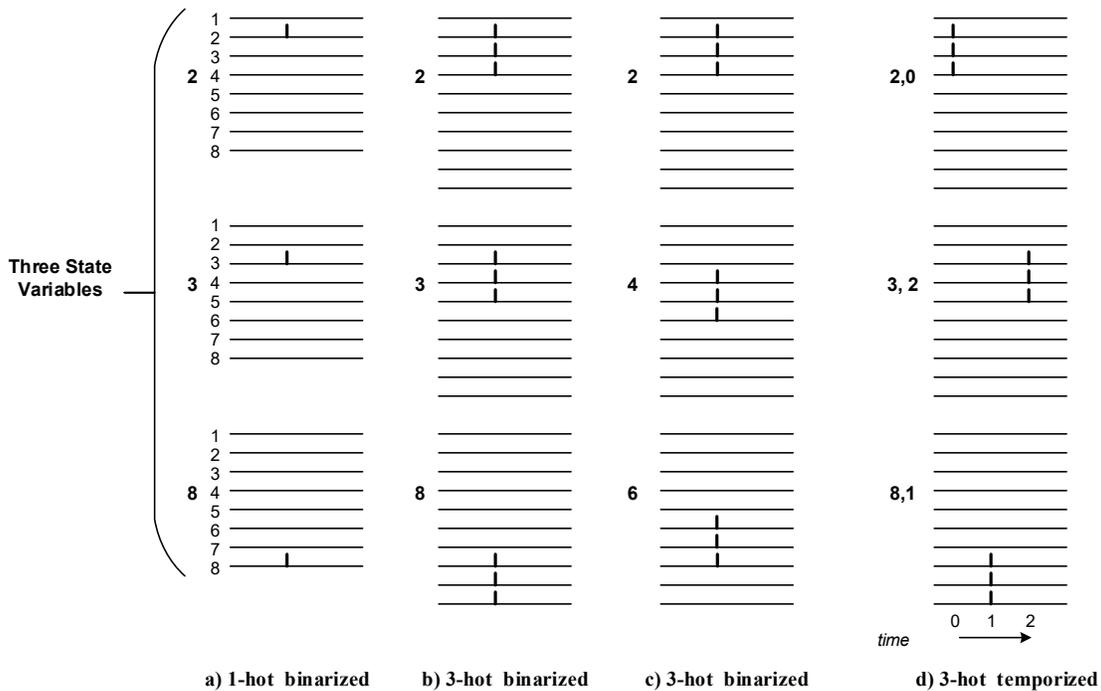

a) 1-hot binarized    b) 3-hot binarized    c) 3-hot binarized    d) 3-hot temporized

**Figure 7 Example mappings of input values to *m*-hot codes. Each spike volley consumes a single time *step*. In this example, each step consists of three time *units* (0, 1, and 2).**

The examples in Figure 7a, b, and c are all *binarized*. There is either a spike or not, and they all occur at the same time. As alluded to above, however, for some applications values may also be *temporized*: the relative time of a spike indicates its relative strength or importance with respect to the other spikes, where earlier spikes are stronger. In Figure 7d, the three values with their associated times are 2,0 | 3,2 | 8,1. Hence, the strongest (or most important) state variable is the first (value = 2) because its spikes occur at time = 0. The weakest is the second state variable (value = 3); its spikes occur 2 time units after the first.

Regarding the cart-pole problem specifically: at discrete intervals, the EN block receives state variables from the environment and produces encoded state variables. In simulation, the environment (EV) operates on 64-bit floating point numbers to model the physics. The EN block reduces the incoming floating point values to small discrete ranges and encodes them as a 3-hot binarized spike volleys.

Note: Two time scales are used in this work. A *step* or *time step* is composed of multiple *time units*. In the cart-pole implementation to follow, a single step is composed of 4 time units. The letter "s" denotes steps, and "t" denotes time units. An exception is the discussion of Q-Learning above in Section 2, where the usual Q-Learning convention is used: "s" represents a state and "t" represents a time step.

### 3.4 C-TNN Architecture

In this section, the basic elements of the C-TNN are described.

*Temporal Communication and Excitatory Neuron Processing*

Communication and processing are illustrated in Figure 8. The mathematics is discrete and operates over the non-negative integers and the "∞" symbol. Spikes occur at points in discrete time, measured according to a unit time clock. The figure shows the processing of a single volley. The first spike in the volley is assigned value (time) 0, and the other spikes occurring at later times are assigned integer values (their times relative to 0). If there is no spike in a given volley, it is assigned "∞".

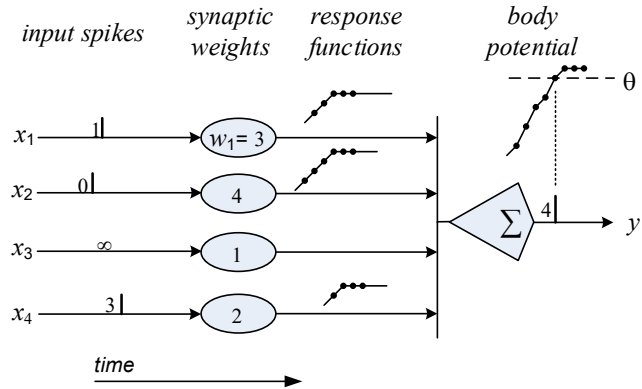

**Figure 8. Example of temporal encoding and excitatory neuron operation. Input values $x_1$ - $x_4$ are communicated as a *volley* of spikes conveyed over a *bundle* of four *lines*. Times increase from left to right, and all times are relative to the first spike observed by the excitatory neuron's integration function (Σ). The input volley encodes values [1, 0, ∞, 3], where ∞ is the assigned value when there is no spike on a given line. As the spikes arrive at the synapses, weights determine the amplitudes of response functions that are passed to the neuron body for summation; the response functions are shifted according to the relative spike times. At the neuron body, the time-shifted response functions are summed to yield the body potential, and an output spike is generated the first time the body potential reaches the threshold θ (at $t$ = 4). If it never reaches the threshold the output spike time is ∞ .**

A useful interpretation of spike time encoding is that the presence of a spike on a line indicates the *presence* of a feature, and the relative spike time indicates the *strength* of that feature. As in the case of biological neurons, spike time resolution is quite low, and the example values in Figure 8 are representative of an actual implementation. As an example, one might interpret a spike at relative time $t$ = 0 as "very strong", at $t$=1 as "medium strong", $t$=2 as "medium weak" and $t$=3 as "very weak".

Excitatory neurons use the spike response model (SRM0)[5]. Figure 8 illustrates SRM0 operation. Synapses in the SRM0 model convert spikes to response functions which are summed, with an output spike being emitted when/if a threshold value is first reached. By choosing different response functions, a designer can influence the neuron's functional capabilities. As depicted in Figure 9, *ramp integrate-and-fire* (RIF) response functions are used in this work. The RIF model has a response function with a sloping leading edge that provides essential temporal processing capabilities.

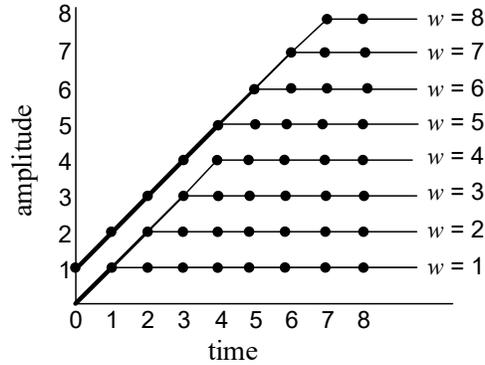

**Figure 9. Discrete ramp response functions for weights 1 through 8. The offset of one time unit for weights 4 and lower accounts for slower body potential rise for lower weights.**

Synaptic learning is achieved via spike timing dependent plasticity (STDP)[3][7], a localized, online method. STDP is described below in the context of columns containing multiple neurons.

*Modeling Inhibition*

Inhibition is modeled as a winner-take-all (WTA) process. A 1-WTA inhibition block has the same number of input and output lines. The input with the earliest spike time is passed through uninhibited as an output spike; all other input spikes yield no output spike.

The handling of tie cases is important because they occur frequently due to the very low precision integers. When there is a 1-WTA tie, the tie breaker first selects spikes emitted from neurons with the highest body potential at the time the spike is generated (so causality is maintained), and any remaining ties may be broken either randomly or systematically. To simplify simulations, ties are broken systematically by selecting the neuron with the lowest index.

*Clustering Columns*

A *column* (Figure 10) implements an online unsupervised clustering function. A sequence of input patterns (spike volleys) is applied to a column's inputs, and the column uses STDP to organize the patterns into clusters: cluster centroids are encoded in the weights. An individual synapse is illustrated in Figure 11.

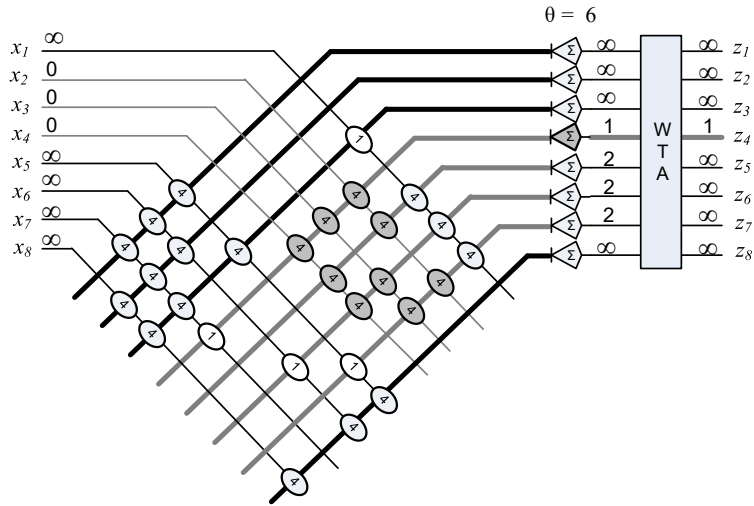

**Figure 10.** A column consists of a synaptic crossbar feeding a set of parallel excitatory RIF neurons. WTA inhibition selects the first output to spike. The crossbar is drawn using an abbreviated notation: a wide line (a bundle) feeding a neuron body contains the individual lines from each synapse. STDP adjusts weights according to input and output spike times. For simplicity, binarized inputs are used in this example, and synapses having 0 weight are not shown. In this example, $w_{max} = 4$, and after sufficient training the synaptic weights form a stable, bimodal distribution. The weights conceptually encode cluster centroids. The fourth neuron (driving $z_4$) has three input synapses at $w_{max}$ that receive spike inputs at $t = 0$. In this case, the body potential of $z_4$ reaches the threshold of 6 at $t = 1$. Three neurons have two synapses of weight $w_{max}$ that receive input spikes, and the summation of their response functions does not reach the threshold until $t = 2$. At the outputs, WTA inhibition selects the fastest spike, so only $z_4$ produces a non-$\infty$ output value. Functionally, WTA's one-hot output volley identifies the cluster centroid nearest to the applied input.

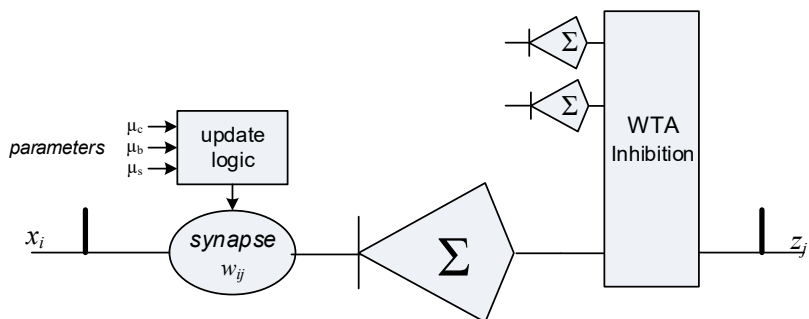

**Figure 11.** The synapse with weight $w_{ij}$ is dependent on timing of spikes on $x_i$ and $z_j$. Update parameters are common to all synapses in the network.

For an applied input pattern, the column outputs a one-hot cluster identifier, determined via the learned centroids. Concurrently, every input pattern in a sequence is assimilated by the learning process so it can be used immediately for clustering later inputs. Because of this continual learning process, over time cluster centroids are free to shift in order to match any drift in the input patterns.

Synaptic weights are potentiated or depressed depending on the spike times of their inputs and the output of their post-synaptic neuron. See Table 2. The first two rows cover the case where both input and output contain a spike, i.e. their spike time is not $\infty$. This is the situation most commonly discussed as classic STDP. If the input spike precedes (or occurs at the same time as) the output spike, then the weight is increased by the value $\mu_c$, up to the maximum weight $w_{max}$. Otherwise, the weight is decreased by $\mu_b$, down to the minimum of 0.

The next three rows cover the remaining cases. If there is an input spike, but no output spike, the weight is increased by $\mu_s$, typically a small value or 0. If there is an output spike, but no input spike, then the weight is decreased by $\mu_b$. Plausibility for the search case may be found in glial cells [12].

**Table 2. Weight update functions for excitatory neurons.**

| input conditions | | weight update | description |
|---|---|---|---|
| $x_i \neq \infty$ | $x_i \leq z_j$ | $\Delta w_{ij} = +\mu_c$ | capture |
| $z_j \neq \infty$ | $x_i > z_j$ | $\Delta w_{ij} = -\mu_b$ | backoff |
| $x_i \neq \infty$ $z_j = \infty$ | | $\Delta w_{ij} = +\mu_s$ | search |
| $x_i = \infty$ $z_j \neq \infty$ | | $\Delta w_{ij} = -\mu_b$ | backoff |
| $x_i = \infty$ $z_j = \infty$ | | $\Delta w_{ij} = 0$ | no-op |

***Example***: Assume a network inputs $x_{1-6}$, outputs $z_{1-4}$, response functions as in Figure 9 with $w_{max} = 8$. All synaptic weights are initialized to 5. Assume all input patterns $x$ consist of 3 spikes, and the neuron threshold is set at 3. Further, assume $\mu_c = 1$, $\mu_b = 1$, and $\mu_s = 0$. When the input volley $x = [\infty\ 0\ \infty\ 0\ 0\ \infty]$ is applied to the column, all the response functions reach the threshold at $t = 0$. All the body potentials are the same, so a systematic 1-WTA tie-break selects the tying neuron with the lowest index, and weights on synapses connected to $z_1$ that receive an input spike are potentiated (+1) and others are depressed (-1). The input pattern has been *captured* by $z_1$, and $z_1$ will start to form a cluster with this pattern as its initial centroid. When a different 3 spike input is applied, $[0\ \infty\ \infty\ 0\ 0\ \infty]$, then $z_1$ will have a body potential below the threshold. All the others tie, and with the systematic tie breaker, $z_2$ captures the second pattern. The example continues through five additional input patterns, where $[\infty\ 0\ \infty\ 0\ 0\ \infty]$ appears twice more. As this happens, the weights of $z_1$'s synapses continue to rise, further strengthening $z_1$'s affinity for the pattern. After the first six inputs, all the neurons have captured a cluster. The 7$^{th}$ input differs from all of the previous. The closest matches are with neurons $z_1$ and $z_2$. The tie-break selects $z_1$. Finally, STDP accounts for the $z_1$ cluster's new member by adjusting $z_1$'s synaptic weights.

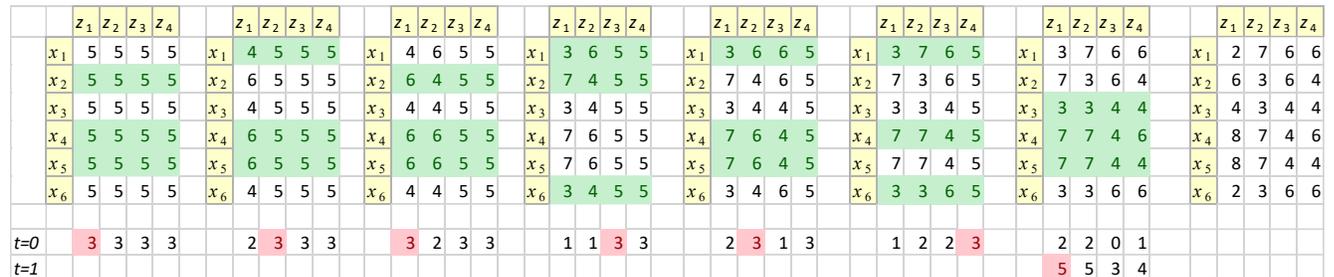

**Figure 12. STDP for initial learning steps. Synaptic weights are shown as a 2-D array above and body potential for times 0 and 1 are shown in the bottom two rows. Green highlights the input pattern during a given step, and red highlights the spiking output for the given input pattern.**

When search mode is enabled ($\mu_s > 0$) new clusters may be captured if there is a macro-level shift in input patterns. If a cluster goes a long time without a cluster member triggering an output spike, search will cause the other, non-cluster, synapses to gradually rise. Eventually, they will trigger a spike and a new cluster is captured.

## 3.5 R-TNN Architecture

The R-TNN performs reinforcement learning. Although inference and learning happen concurrently, they are better explained separately.

Referring to Figure 4, an R-TNN column has $p$ 1-hot inputs $x_1..x_p$, that correspond to CIds coming from the C-TNN. It has $q$ 1-hot outputs $z_{1..q}$, which correspond to the types of output actions. Thus, the column contains a $p \times q$ synaptic crossbar coupling its inputs to the $q$ neuron bodies which generate output spikes.

Synaptic learning in the R-TNN employs neo-Hebbian three-factor learning rules as put forward by Gerstner et al. [4]. The first two factors are the same as with conventional two-factor STDP: a synapse's input spike time and its post-synaptic neuron's output spike time. The third factor is a broadcast reward signal R (a punishment is a negative reward).

Because the system acts in a series as steps, the one-hot input at step $s$ is denoted $x_i(s)$, the one-hot output is $z_j(s)$, and the reward signal is R($s$). Reminder: a single *step* consists of multiple *time units*.

*Weight Update* **Implementation.**

Just as with conventional STDP, each synapse has its own local update logic, the only shared signal is the broadcast reward. In general, there may be both rewards and punishments in the same system, and this is the case with the cart-pole implementation described here. The three factor weight update function is shown schematically in Figure 13. The three factors are the $x_i$, the $z_j$, and the broadcast reward R.

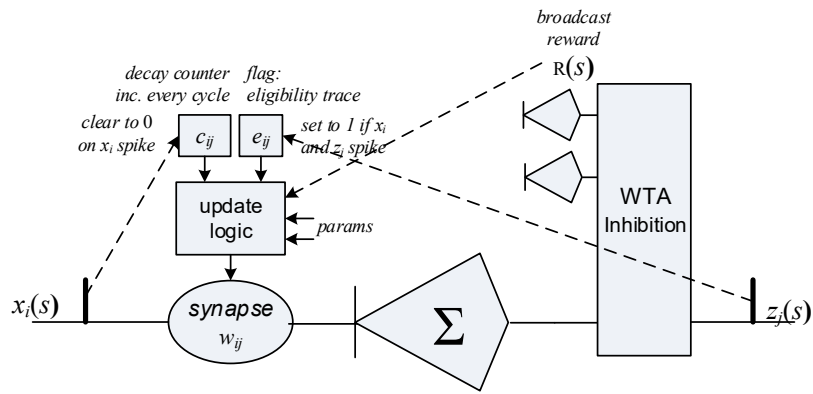

**Figure 13. The R-TNN synapse implementation employs a neo-Hebbian three factor update rule.**

A key part of the update process is a time window $\omega$ that contains the most recent spiking outputs $z_j$, paired with the $x_i$ that triggered the spike. When a reward is broadcast, all the synapses belonging to all the neurons in the window are subject to three-factor STDP update.

As shown in Figure 13, the implementation of synapse $i,j$ consists of:

*Variables / State:*

    $w_{ij}$ : a weight up/down counter that saturates at 0 and $w_{max}$. Weights are initialized at an initial value $w_{init}$.

    $c_{ij}$: a counter that is cleared to 0 whenever a spike on $x_i$ is received. At each subsequent step it counts up until it saturates at $\omega_\rho$ or $\omega_\pi$.

    $e_{ij}$: a binary flag that sets to 1 if output $z_j$ spikes during a step that $x_i$ spikes. It is cleared to 0 if $z_j$ does not spike during a step that $x_i$ spikes.

    R: the reward signal; +1 if there is a reward, -1 if there is a punishment

*Reward (ρ) Parameters:*
    $\omega_\rho$: update window size
    $\rho_0^+$: maximum potentiation amount (when $c_{ij} = 0$).
    $\rho_0^-$: maximum depression amount (when $c_{ij} = 0$).

*Punishment (π) Parameters*
    $\omega_\pi$: update window size
    $\pi_0^+$: maximum potentiation amount (when $c_{ij} = 0$).
    $\pi_0^-$: maximum potentiation amount (when $c_{ij} = 0$).

The potentiation and depression values decay over the length of the window. Assuming linear decay, for synapse *i,j*, update values for $w_{ij}$ are:

$$\rho_{ij}^+ = \rho_0^+ - c_{ij} * \rho_0^+ / \omega_\rho$$
$$\rho_{ij}^- = \rho_0^- - c_{ij} * \rho_0^- / \omega_\rho$$
$$\pi_{ij}^+ = \pi_0^+ - c_{ij} * \pi_0^+ / \omega_\pi$$
$$\pi_{ij}^- = \pi_0^- - c_{ij} * \pi_0^- / \omega_\pi$$

Because the output neurons form a 1-hot code, at most one of the neurons will spike. When a reward is received at step *s*, all synaptic weights for neurons along the action path are updated, not just synapses that received an input spike (as determined by $e_{ij}$). For any synapse, *i,j*, if $c_{ij} < \omega$ then the associated synaptic weight should be updated according to the following update rules:

    If $R(s) = 1$ then
        if $e_{ij} = 1$ then $\Delta w_{ij} = \rho_{ij}^+$
        if $e_{ij} = 0$ then $\Delta w_{ij} = \rho_{ij}^-$
    else if $R(s) = -1$ then
        if $e_{ij} = 0$ then $\Delta w_{ij} = \pi_{ij}^+$
        if $e_{ij} = 1$ then $\Delta w_{ij} = \pi_{ij}^-$

For an example, refer to Figure 14.

Note that the same update parameters apply to all neurons and synapses within a column. They are established at the time the system is initialized, and they are not affected by the learning process. Also note that with the exception of the broadcast reward signal, synaptic update rules act on local information: whether the synapses's input and/or its neuron's output has spiked.

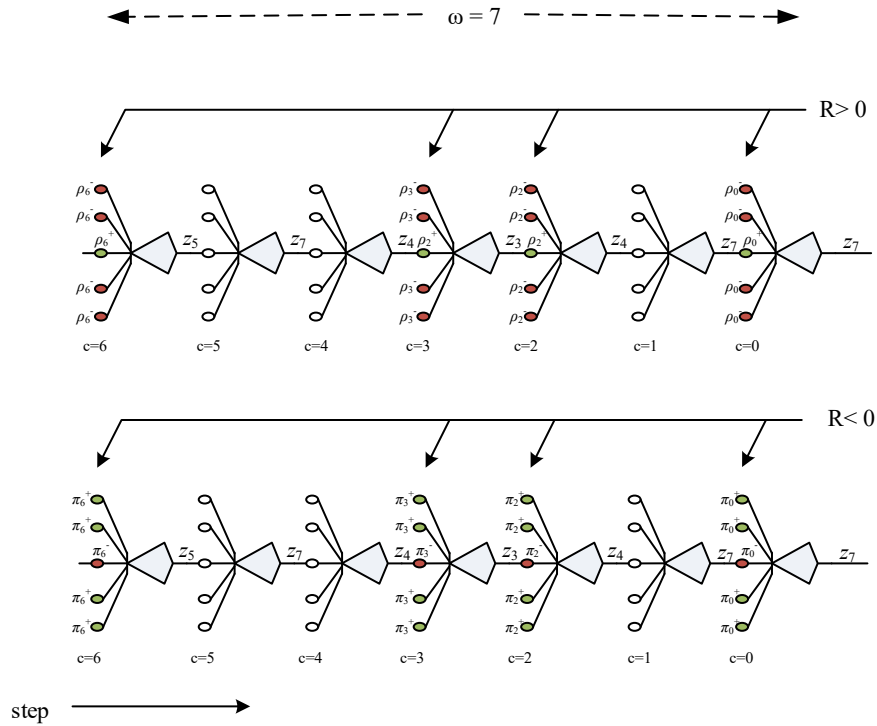

**Figure 14. Three factor synaptic updates. A dynamic (unrolled) sequence of steps along an action path is shown; actions are encoded as neuron spikes $z_i$. The same neuron may appear more than once, but only the most recent dynamic occurrence results in a synaptic update. All synapses associated with a neuron on the path are subject to updates. If the reward is positive (R > 0), the synapses directly on the path are potentiated and other synapses belonging to the same neuron are depressed. If the reward is negative, synapses directly on the path are depressed and the others are potentiated. The amounts of potentiation and depression decrease along the path of length ω, as determined by counter values (*c*).**

*R-TNN Inference*

Inference in an R-TNN is a simple version of inference in a regular TNN. An example trained R-TNN column is in Figure 15. This is a 1-hot to 1-hot mapping, so inference basically determines the highest weighted output for the given spiking input. There is no need to add synaptic weights because there can be most one active synapse per neuron.

The excitatory neurons have a threshold $θ_R$ that is kept relatively low so in most cases there is an output spike. If there is a tie, then the first tie breaker chooses the neuron with the highest body potential (i.e. a single weight). The second tie breaker is a *consistent* pseudo-random selection. It is consistent in the sense that once a tie has been decided in a given way, the same decision continues to be used for following consecutive ties. Finally, if the threshold $θ_R$ is not reached by any of the neurons, then a consistently pseudo-random neuron is selected to spike.

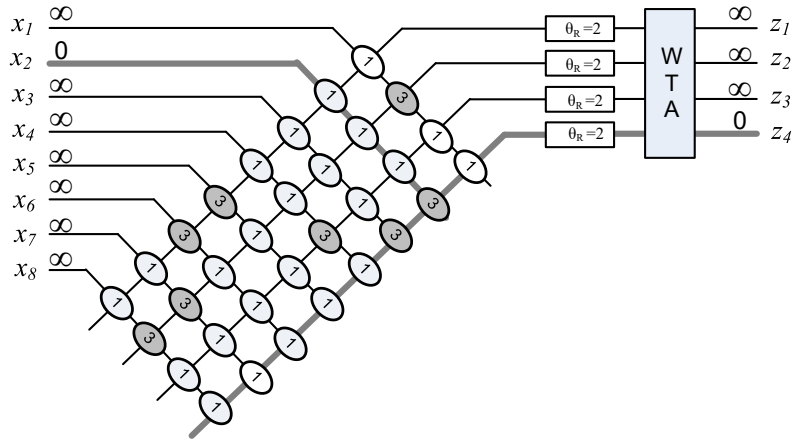

**Figure 15. R-TNN column.** $w_{max}$ = 3. The outputs $z_i$ indicate actions. For the cart-pole problem, there are only two actions that correspond to application of forces +F and -F.

## 4. Initial Simulation Study

To illustrate the basic design methodology, a simple system is first simulated. Two TNNs each consisting of a single column cooperate to balance the pole. The frontend C-TNN feeds the backend R-TNN, and there is a feedback path from the R-TNN to the C-TNN through the environment.

For initial simulations only one state variable is used: the pole angle. The pole angle is restricted to fall within a range of ± 12º, and this range is discretized into 16 equal intervals and encoded in a 3-hot code (Figure 7). Codings are binarized (Figure 7b).

As a matter of methodology, the first set of simulations are directed at the R-TNN back-end rather than the C-TNN clustering front-end. After the R-TNN has been studied and optimized, a subsequent set of simulations focuses on the C-TNN.

*Episodes*

A simulation run consists of multiple episodes, each beginning with a random initial pole angle: -2.0 ≤ ∠ ≤ +2.0 and initial displacement: $d$ = 0. Synaptic weights (and therefore learning) are preserved between episodes. Simulation of each episode proceeds as a series of steps until there is a failure (∠ falls out of range ± 12º or $d$ exceeds ± 2.4 meters) or until 10,000 steps are reached successfully (i.e., 200 seconds of simulated time).

A pole failure yields negative reward (R($s$) = -1). In the initial simulations, the track displacement $d$ is not used as a state variable, so a track failure results in zero reward (R($s$) = 0). Nevertheless, a track failure terminates an episode. If a simulation reaches at least 500 steps (and for every 500 steps thereafter), a reward R($s$) = +1 is generated. In all other cases R($s$) = 0.

For the dimensions and parameters used here, if there is a failure, it is because the cart hits the end of the track. The common scenario is that the pole is kept within range for long periods of time, but there is a slow drift in cart displacement (Figure 16). Eventually, the cart drifts to the end of the track. Of course, this behavior is to be expected because displacement is not used as a state variable.

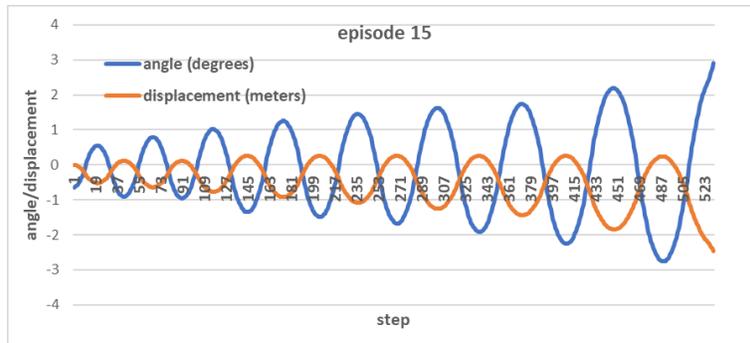

**Figure 16. Pole angle and cart displacement as a function of the step. Results for a typical simulation run are shown. The cart gradually drifts to the left until it hits the end of the track. Meanwhile, the pole angle remains within valid range.**

*Metrics*

The primary metric is based on numbers of successful steps each episode consumes, i.e., how long the pole remains upright, and the cart is on-track.

To compute the metric, the network is initially warmed up for a number of "training" episodes, and the simulation continues for a number of "test" episodes. Training continues as an ongoing process during the test phase. After all the episodes are complete, the numbers of steps per episode are averaged over the number of test episodes.

*Random Number Seeds*

The initial pole angle for each episode is generated by a Matlab pseudo-random number generator. It was found that the specific seed has a significant effect on the results because different seeds lead to different warm-up sequences. To avoid inadvertent cherry-picking by using what happens to be a good seed, eight random seeds -- 1 through 8 -- are used, and simulation results for all 8 are given (or are averaged).

*Naive/Optimal Approach*

An ideal method implemented as a standard for comparison. With only one state variable, if the pole leans left, then apply force to the left, and if the pole leans right, apply force to the right. In the absence of any other information, this naive approach is also optimal in a sense -- given the sparse state information (only the pole angle) it appears to be the only viable approach.

*Q-Learning*

A Q-Learning method was implemented as a standard for comparison. The method uses the TNN for initial exploration and Q-Table updates, and then an exploitation test phase uses the Q-Table to determine actions in the usual way (i.e., by selecting the action with the highest value in a given row). Q-Learning continues during the test phase. The implementation is a text-book implementation using Bellman's equation. The γ and α parameters were determined via good-faith, manual design space exploration; there are no guarantees of optimality.

With this method, at the time the exploration phase is turned on, both methods have observed exactly the same sequence of input state variables and have had the opportunity to update weights or table entries. Comparisons are then fair in the sense that both methods have explored the same state space when the test phase begins.

## 4.1 Simulation Results

*Naive/Optimal*

When the only state variable is the angle, the only reasonable strategy is to push the cart to the right when the pole leans to right and to push the cart to the left when the pole leans to the left. Accordingly, weights in the R-TNN were manually set at fixed values to achieve this. The balancing methods were simulated for 8 different pseudo-random generator seeds. Then the average number of successful steps during the test phase were sorted from best to worst. This is to give an idea of the variability due to the individual *png* seeds.

For comparison, results for this manually weighted method and the Q-Learning and T-learning methods for 30+50 (warmup + test) are in Figure 17. Allowing for variations due to pseudo-random seeds, the T-learning method is on par with the optimal method. Given that the pole angle is the only input, all three methods appear to be at the performance ceiling.

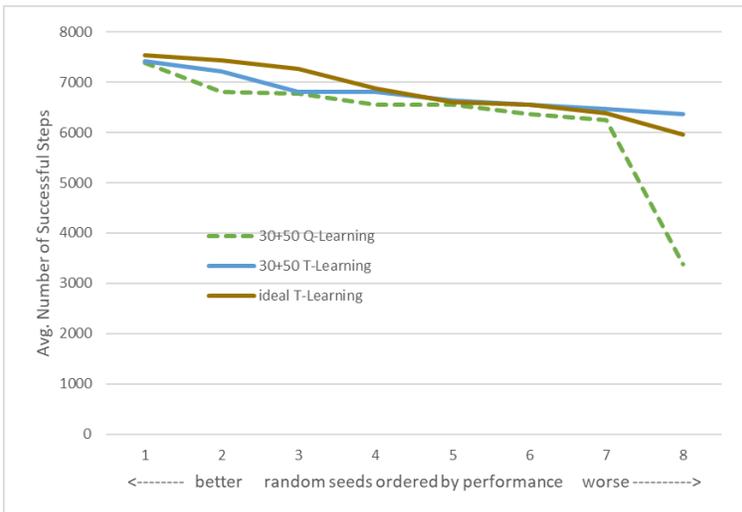

**Figure 17. For a single state variable (the angle), Ideal, 30+50 Q-Learning, and T-Learning yield similar performance except for one of the Q-Learning runs.**

*R-TNN Implementation*

R-TNN The initial baseline system uses binarized 3-hot inputs (Figure 7b and c). In both the C-TNN and R-TNN, the maximum weight $w_{max} = 8$ and weights are initialized to $w_{init} = 5$. Other parameters are listed in Table 3. These parameter values were determined via a series of initial exploratory simulations. Some of the weight increments and decrements are fractions, so this adds a few bits of precision to the weight counters. However, the ceiling function is applied to weights before they used for inference, so inference is a purely small integer operation.

Table 3. Column Configurations for Initial Simulations

| C-TNN parameter | value | description |
|---|---|---|
| $\theta_C$ | 6 | threshold |
| $\mu_c$ | 1/16 | capture |
| $\mu_b$ | 1/16 | backoff |
| $\mu_s$ | 0 | search |
| **R-TNN parameter** | **value** | **description** |
| $\theta_R$ | 2 | threshold |
| $\rho_0^+, \rho_0^-$ | 3/2 | reward potentiation and depression |
| $\omega_\rho$ | 2 | reward window |
| $\pi_0^+, \pi_0^-$ | 3/2 | punishment potentiation and depression |
| $\omega_\pi$ | 16 | punishment window |

For comparison Q-Learning was used. For Q-Learning, γ = .95 and α = .9. Each simulation was run for a variable number warm-up (training) episodes, from 10 to 50, each followed by 50 test episodes. The performance metric is the average number of successful steps per episode. As noted above, due to variations caused by the random number seed, 8 different seeds are simulated. Then, the 8 results are each sorted from worst to best, irrespective of the seeds, and plotted. Simulation results are in Figure 18.

The Q-Learning method is capable of slightly higher performance for some seeds, but also suffers significantly lower performance for other random seeds. An explanation is that the T-Learning method learns so fast that the Q-Table exploration phase is not long enough to thoroughly explore the state space. Some outlier states are seldom or never visited, and this leads to incomplete Q-Learning. It appears that the 30+50 case works best for Q-Learning; in that case only one of the 8 random seeds results in a fall-off in performance. It seems likely, however, that if one were to use a more sophisticated exploration strategy, the Q-Learning method would be on par with the T-Learning method across the board.

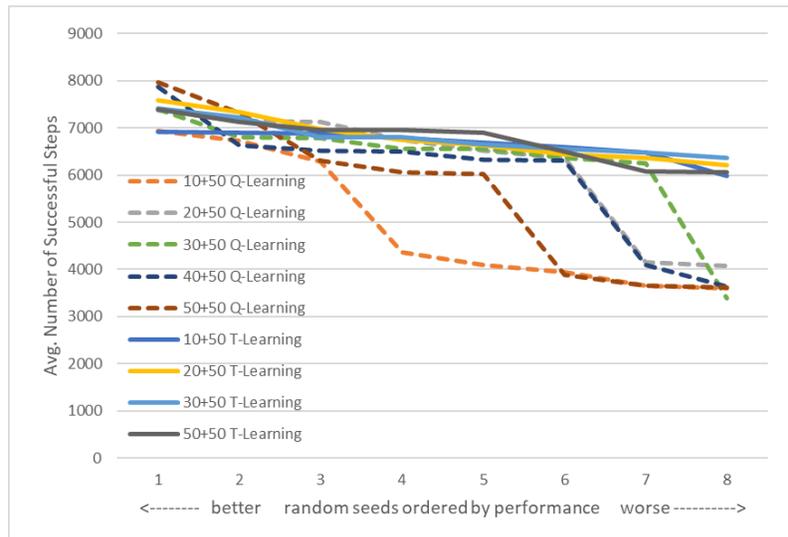

**Figure 18. Average number of successful steps per test episode for varying number of training episodes. The *x*-axis represents 8 different random number generator seeds. Each curve reflects sorted performance from worst to best for a given simulation, irrespective of the specific seeds.**

## 4.2 C-TNN Design

In previous simulations, the number of CIds (and excitatory model neurons) was the same as the number of input states (16), so no true clustering was performed. To exploit true clustering, a set of simulations with reduced numbers of C-TNN neurons were run.

Parameters are the same as in Table 3, except that search mode is used ($\mu_s$ = 1/128). Simulation results are plotted in Figure 19. The number of neurons can be cut in half with no significant reduction in performance. Reducing the neurons to 6 results in reduction for two of the random seeds.

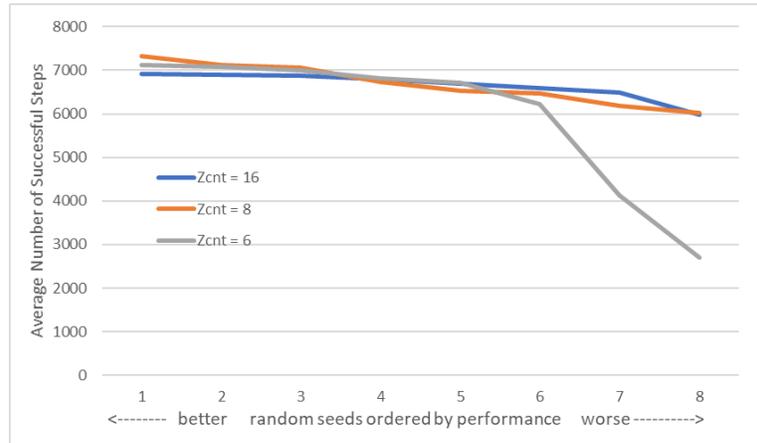

**Figure 19. Performance for reduced numbers of C-TNN neurons (Zcnt).**

*Summary*

The combination of C-TNN and R-TNN begin with a blank slate of synaptic weights. The pole angle is streamed in at the rate of one sample per 20 msec., expressed as one of 16 intervals. The C-TNN learns a mapping that reduces the 16 intervals onto 8 clusters, based on similarity. Then, the 8 CIds are streamed to the R-TNN that learns, online, the optimal action for each of the clusters. The action, when applied to the cart, affects the pole angle which is fed back into the C-TNN. Although the problem is a relatively simple one, all the components of the RL system work as expected.

## 5. Multiple State Variables

Although the single state variable system just described seems to perform reasonably well, many episodes fail because the cart eventually drifts to an end of the track (Figure 16).

To improve performance, state variable(s) can be added to allow the agent to adjust for the drift. To this end, four state variables and intervals are taken from [10], using the lower precision "15 bins" intervals given in that work. The ranges and intervals for each of the four state variables are in Table 4. The interval# is used in Figure 20 for indexing synaptic weights. One hot encodings are also given in the table.

As an aside, it is generally better to use the least precision that will give satisfactory results. In a sense, reducing precision is a simple way of performing similarity clustering.

**Table 4. State Variable Intervals and Encodings**

|  | interval# | interval | encoding |
|---|---|---|---|
| angle ∠ | 1 | -12, -6 | 100000 |
|  | 2 | -6, -1 | 010000 |
|  | 3 | -1, 0 | 001000 |
|  | 4 | 0, 1 | 000100 |
|  | 5 | 1, 6 | 000010 |
|  | 6 | 6, 12 | 000001 |
| displacement *d* | 1 | -2.4, -.8 | 100 |
|  | 2 | -.8, .8 | 010 |
|  | 3 | .8, 2.4 | 001 |
| angular velocity d∠/d*t* | 1 | *-inf*, -50 | 100 |
|  | 2 | -50, +50 | 010 |
|  | 3 | 50, *+inf* | 001 |
| cart velocity d*d*/d*t* | 1 | *-inf.*, -5 | 100 |
|  | 2 | -5, +5 | 010 |
|  | 3 | 5, *+inf* | 001 |

In the simulations in this section, all episodes begin with a pseudo-randomly chosen initial angle between plus and minus 1.5 degrees. The metric is the same as used earlier: it is the number of successful steps averaged over a set of consecutive episodes (typically 30), with the maximum of 10,000.

During initial simulations, it became evident that the angular velocity of the pole is always in the mid-range. In other words it provides no useful information and is not considered further. This leaves three state variables, and the following three systems are considered:

       1 SV *system*      angle only

       2 SV *system*      angle & cart velocity

       3 SV *system*      angle & cart velocity & cart displacement

*Optimal Performance*

For the next step, through an *ad hoc* method, weights yielding what is apparently optimal performance were determined and were coded into the synapses manually. These will be referred to as "optimal" for lack of a better term, although they have not been strictly proven to be optimal. See Figure 20. Optimal synaptic weights -- strongly conjectured to be optimal, but not proved to be so. Ranges for each of the state variable ranges can be found in Table 4. The 1 SV weights lead to behavior as described above: if the pole leans to the left, then force is applied to push the cart to the left (-F). If the pole leans to the right, a force of +F is applied. For 2 SV and 3 SV systems, the weight matrix is symmetric around the pole angle and is more complex but does make intuitive sense when examined closely.

### a) 1 SV system

| angle | -F | +F |
|---|---|---|
| 1 | 8 | 0 |
| 2 | 8 | 0 |
| 3 | 8 | 0 |
| 4 | 0 | 8 |
| 5 | 0 | 8 |
| 6 | 0 | 8 |

### b) 2 SV system

| angle | d$d$/d$t$ | -F | +F |
|---|---|---|---|
| 1 | 1 | 0 | 8 |
| 1 | 2 | 8 | 0 |
| 1 | 3 | 8 | 0 |
| 2 | 1 | 0 | 8 |
| 2 | 2 | 8 | 0 |
| 2 | 3 | 8 | 0 |
| 3 | 1 | 8 | 0 |
| 3 | 2 | 0 | 8 |
| 3 | 3 | 8 | 0 |
| 4 | 1 | 0 | 8 |
| 4 | 2 | 8 | 0 |
| 4 | 3 | 0 | 8 |
| 5 | 1 | 0 | 8 |
| 5 | 2 | 0 | 8 |
| 5 | 3 | 8 | 0 |
| 6 | 1 | 0 | 8 |
| 6 | 2 | 0 | 8 |
| 6 | 3 | 8 | 0 |

### c) 3 SV system

| angle | d | d$d$/d$t$ | -F | +F |
|---|---|---|---|---|
| 1 | 1 | 1 | 8 | 0 |
| 1 | 1 | 2 | 0 | 8 |
| 1 | 1 | 3 | 8 | 0 |
| 1 | 2 | 1 | 8 | 0 |
| 1 | 2 | 2 | 8 | 0 |
| 1 | 2 | 3 | 8 | 0 |
| 2 | 1 | 1 | 8 | 0 |
| 2 | 1 | 2 | 0 | 8 |
| 2 | 1 | 3 | 8 | 0 |
| 2 | 2 | 1 | 8 | 0 |
| 2 | 2 | 2 | 8 | 0 |
| 2 | 2 | 3 | 8 | 0 |
| 2 | 3 | 1 | 0 | 8 |
| 2 | 3 | 2 | 8 | 0 |
| 2 | 3 | 3 | 8 | 0 |
| 3 | 1 | 1 | 0 | 8 |
| 3 | 1 | 2 | 0 | 8 |
| 3 | 1 | 3 | 8 | 0 |
| 3 | 2 | 1 | 8 | 0 |
| 3 | 2 | 2 | 0 | 8 |
| 3 | 2 | 3 | 8 | 0 |
| 3 | 3 | 1 | 8 | 0 |
| 3 | 3 | 2 | 0 | 8 |
| 3 | 3 | 3 | 8 | 0 |
| 4 | 1 | 1 | 0 | 8 |
| 4 | 1 | 2 | 8 | 0 |
| 4 | 1 | 3 | 0 | 8 |
| 4 | 2 | 1 | 0 | 8 |
| 4 | 2 | 2 | 8 | 0 |
| 4 | 2 | 3 | 0 | 8 |
| 4 | 3 | 1 | 0 | 8 |
| 4 | 3 | 2 | 8 | 0 |
| 4 | 3 | 3 | 8 | 0 |
| 5 | 1 | 1 | 0 | 8 |
| 5 | 1 | 2 | 0 | 8 |
| 5 | 1 | 3 | 8 | 0 |
| 5 | 2 | 1 | 0 | 8 |
| 5 | 2 | 2 | 0 | 8 |
| 5 | 2 | 3 | 0 | 8 |
| 5 | 3 | 1 | 0 | 8 |
| 5 | 3 | 2 | 8 | 0 |
| 5 | 3 | 3 | 0 | 8 |
| 6 | 2 | 1 | 0 | 8 |
| 6 | 2 | 2 | 0 | 8 |
| 6 | 2 | 3 | 0 | 8 |
| 6 | 3 | 1 | 0 | 8 |
| 6 | 3 | 2 | 8 | 0 |
| 6 | 3 | 3 | 0 | 8 |

**Figure 20. Optimal synaptic weights -- strongly conjectured to be optimal, but not proved to be so. Ranges for each of the state variable ranges can be found in Table 4.**

For each of the optimal systems, a total of 32 trials with random initial angles and random weights were simulated and results are plotted in Figure 21. The optimal 1 SV system performs reasonably well (given that episodes top out at 10,000 steps), however, adding the cart velocity in the 2 SV system improves performance significantly, and the 3 SV system is better yet with one of the trials achieving 30 consecutive episodes that all max out at 10,000 steps.

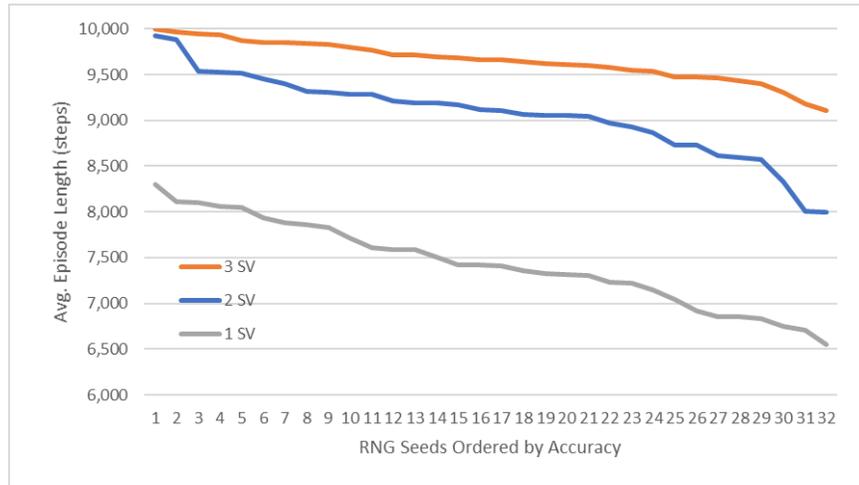

**Figure 21. Optimal accuracies for 32 trials with pseudo-random initial angles and weights.**

## 5.1 2 SV Systems

**Table 5. Column Configurations for 2 SV Simulations**

| C-TNN parameter | value | description |
|---|---|---|
| $\theta_C$ | 12 | *threshold* |
| $\mu_c$ | 1/16 | *capture* |
| $\mu_b$ | 1/16 | *backoff* |
| $\mu_s$ | 0 | *search* |
| **R-TNN parameter** | **value** | **description** |
| $\theta_R$ | 2 | *threshold* |
| $\rho_0^+, \rho_0^-$ | 1 | *reward potentiation and depression* |
| $\omega_\rho$ | 1 | *reward window* |
| $\pi_0^+, \pi_0^-$ | 1 | *punishment potentiation and depression* |
| $\omega_\pi$ | 32 | *punishment window* |

After some performance tuning (see Table 5), a set of simulations for the 2 SV system was run, using 32 pseudo-random seeds. Each of the 32 trials consists of warm-up sequence of 170 episodes, followed by 30 episodes where the number successful steps are recorded and averaged (see Figure 22).

For some of the trials the synaptic weights do not converge (right side of the figure). When more warmup episodes are added, the weights eventually converge. In the middle region of Figure 22, the weights converge, but they converge to the 1 SV weights, i.e., there is no dependence on the cart velocity. Finally at the extreme left end of the plot are the trials where weights converge to the 2 SV optima.

With a single state variable 1 SV, convergence to the optimum is straightforward so the naive approach is quickly learned. By adding state variables, however, the problem becomes a non-linear optimization problem where there are multiple stable sets of weights. Many of these stable states lead to the 1 SV optimum. Only a few lead to the 2 SV optimum (and there are hybrids in between).

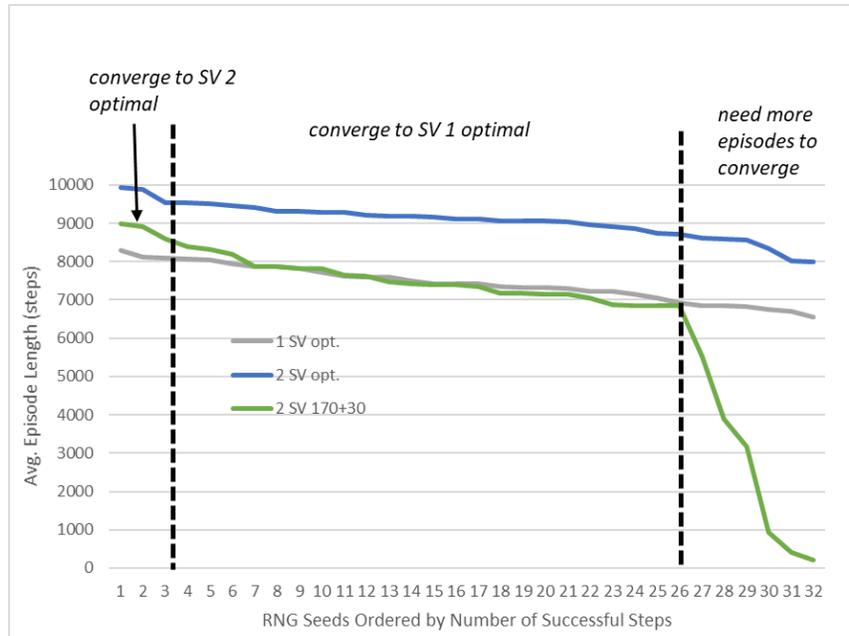

**Figure 22. Performance (average episode length) for the 2 SV system, with optimal 1 SV and 2 SV performance given for comparison.**

Hence, there are multiple sets of stable weights (and multiple local optima). Each random seed sends the system toward some stable state.

Regarding rate of convergence, 16 trials with pseudo-random seeds were simulated, where convergence of some type is assumed when a moving average of at least 6000 successful steps over 30 episodes was achieved. That is, at least 1 SV convergence is reached. Results are in Figure 23. Keep in mind, that the 30 measured episodes are included in the totals, so, for example, the trial that "converged" at around 50 episodes, actually converges after 50-30 = 20 episodes.

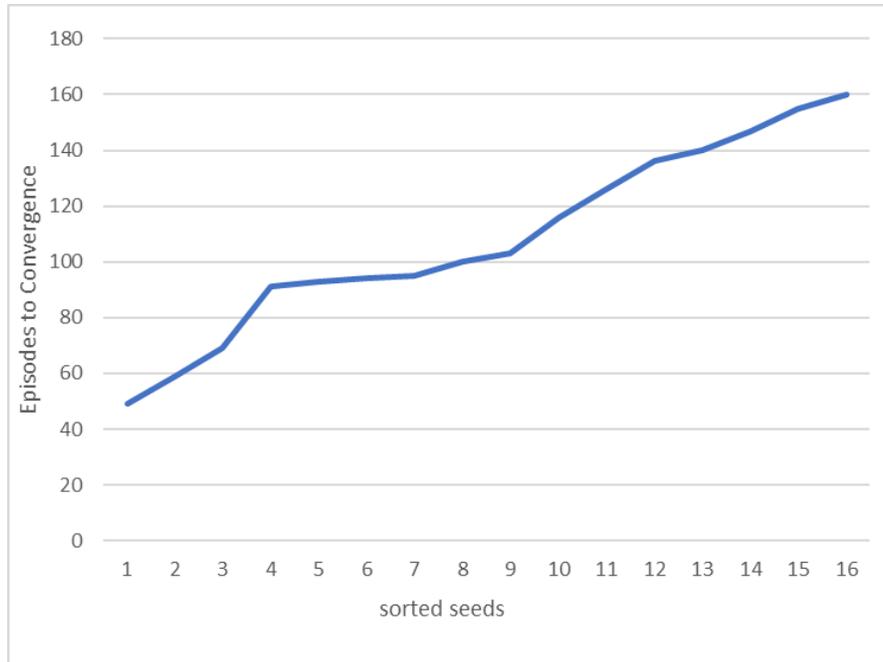

**Figure 23. Numbers of episodes to reach an average of 6,000 successful steps over a sliding window of 30 consecutive episodes.**

Finally, the angle and cart displacement for one of the simulations that achieved the 2 SV optimum is in Figure 24. This solution holds the angle to a very tight range of about + or - 1 degrees. Meanwhile, the cart displacement is kept within range through a process of constant adjustments. This is the type of improved behavior that adding SVs was intended to achieve.

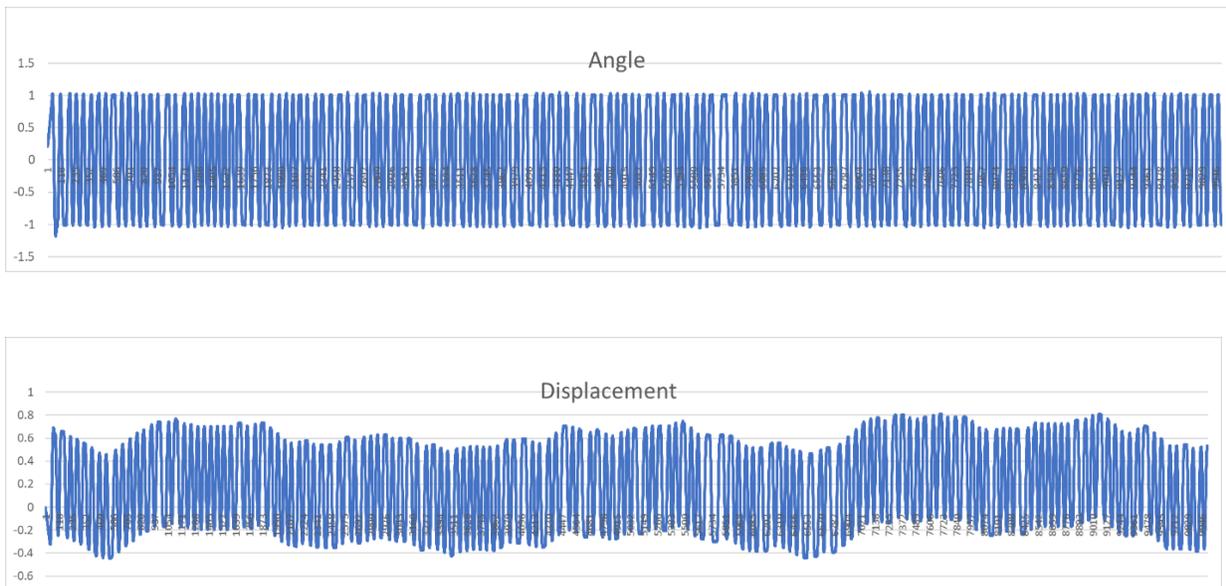

**Figure 24. Simulation results for a 2 SV optimum system.**

*Discussion and Follow-on Research*

This is ongoing research. It appears 3-factor STDP enables relatively quick convergence to a stable set of weights. So, the basic mechanism drives the weights toward stability, and in some cases it drives the system toward the 2 SV optimum. However, in the majority of cases it drives the system toward non-optimum 1 SV stability. The challenge, then, is the same as in many non-linear optimization problems: to somehow steer the system toward a global optimum and away from non-optimal local stability.

We are faced with what amounts to the stability plasticity dilemma. A general approach is to add some randomness to the convergence process.

A sledge-hammer approach employs an agent that starts with pseudo-random weights and runs for some specified number of episodes. If the target performance is not achieved (say 30 consecutive episodes with 9,000 average steps), then the agent re-sets the weights to a new set of pseudo-random values and the process can be repeated. Eventually (see the left side of Figure 22) the objective will be achieved, i.e., convergence to the optimum 2 SV weights. Although this method will work, it is not a very satisfying solution.

Follow-on research will consider more subtle ways of adding randomness to the convergence process, say by periodically re-randomizing a small subset of the weights.

Other follow-on research will use the previous action(s) as an additional input to the process. This will lead to networks that are not only recurrent through the environment, but also internally recurrent.

## 6. Concluding Remarks

The *column* appears to be a fundamental TNN building block; it is composed of 1) a set of parallel excitatory neurons fed by 2) a synaptic crossbar and feeding into 3) a 1-WTA inhibition block. To draw an analogy: in conventional logic design terms, the column is a universal RTL building block. Its function is fairly powerful, like an ALU or mux, and it easy to understand and apply. Multiple columns are connected so they can work in concert to perform some cognitive task. The column has several variations, two of which are demonstrated in this paper. Both the C-TNN and R-TNN are composed of columns: they follow the same biologically plausible paradigm, although they are configured differently.

The frontend/backend architecture may be archetypical for a wide variety of RL systems. In an agnostic way, the frontend merely looks for similarities in the inputs it observes. Then, the RL backend uses the similarity information to determine the action to be taken for a given input. The author has found this architecture and division of labor for other RL applications.

Finally, regarding reinforcement learning: the neo-Hebbian update rules articulated by Gerstner et al. [4] are demonstrated to work in a biologically plausible spiking neuron system performing a reinforcement learning task.